\documentclass{article}
\usepackage[final]{neurips_2026}
\makeatletter
\renewcommand{\@noticestring}{}
\makeatother
\usepackage{comment}
\usepackage{graphicx}
\usepackage{multirow}
\usepackage[utf8]{inputenc} 
\usepackage[T1]{fontenc}    
\usepackage{hyperref}       
\usepackage{url}            
\usepackage{booktabs}       
\usepackage{amsfonts}       
\usepackage{nicefrac}       
\usepackage{microtype}      
\usepackage{xcolor}         
\usepackage{bbding} 
\usepackage{pifont} 
\usepackage{subcaption} 
\title{Multi-domain Multi-modal Document Classification Benchmark with a Multi-level Taxonomy}

\author{%
\textbf{Denghao Ma}$^{1}$ \quad \textbf{Qing Liu}$^{2}$ \quad \textbf{Zulong Chen}$^{2}$\thanks{indicates project leader.} \quad \textbf{Chuanfei Xu}$^{3}$ \\
\textbf{Jia Xu}$^{4}$ \quad \textbf{Wei Shao}$^{2}$ \quad \textbf{Zhibo Yang}$^{2}$ \quad \textbf{Zhao Li}$^{5}$ \\[0.5em]
$^{1}$Beijing Information Science and Technology University, China \quad $^{2}$Alibaba Group, China \\
$^{3}$Guangdong Laboratory of Artificial Intelligence and Digital Economy (SZ), China \\
$^{4}$Guangzhou University, China \quad $^{5}$Zhejiang Lab, China
}

\begin{document}

\maketitle

\vspace{-2mm}
\begin{abstract}
Document classification forms the backbone of modern enterprise content management, yet existing benchmarks remain trapped in oversimplified paradigms— single domain settings with flat label structures—that bear little resemblance to the hierarchical, multi-modal, and cross-domain nature of real-world business documents. This gap not only misrepresents practical complexity but also stifles progress toward industrially viable document intelligence. To bridge this gap, we construct the first \underline{M}ulti-level, \underline{M}ulti-domain, \underline{M}ulti-modal document classification \underline{Bench}mark (\textbf{MMM-Bench}). MMM-Bench includes (1) a deeply hierarchical taxonomy spanning five levels that capture the authentic organizational logic of business documentation; and (2) $5,990$ real-world multi-modal documents meticulously curated from $12$ commercial domains in Alibaba. Each document is manually annotated with a complete hierarchical path by domain experts. We establish comprehensive baselines on MMM-Bench, which consists of open-weight models and API-based models. Through systematic experiments, we identify four fundamental challenges within MMM-Bench and propose corresponding insights. To provide a solid foundation for advancing research in multi-level, multi-domain document classification, we release all of the data and the evaluation toolkit of MMM-Bench at \url{https://github.com/MMMDC-Bench/MMMDC-Bench}.
\end{abstract}

\section{Introduction}
Document classification plays a key role in modern enterprise operations~\citep{NAKAYAMA2024102717}, enabling a wide range of applications, such as automated document archiving~\citep{DBLP:journals/tip/GerekCTA99}, intelligent information retrieval~\citep{DBLP:conf/sigir/Zarri84}, and workflow automation~\citep{DBLP:conf/sigmod/Sheth95}. However, the characteristics of real-world business documents make automatic classification particularly challenging. First, they are \textbf{hierarchically structured}: a document typically belongs to multiple nested categories, for instance, an invoice may fall under ``financial documents'' at the top level, and ``utility bills'' at the fine-grained level. Second, they are \textbf{multi-modal} in nature, integrating diverse content types such as natural language text, tabular data, logos, signatures, and complex visual layouts. Third, they are \textbf{multi-domain}, often carrying relevance across multiple domains.

Despite recent advances in hierarchical classification~\citep{DBLP:conf/nips/GorenGE24,DBLP:conf/nips/RenS023} and multi-modal document understanding~\citep{DBLP:conf/acl/BaiLMLLLXZWHWJL23,DBLP:conf/acl/TuGCT23}, no existing benchmark jointly incorporates hierarchical taxonomies, multi-modal content, and multi-domain scenarios within a single corpus. For example, \textit{RVL-CDIP}~\citep{NEURIPS2022_4c0986bd} provides document images but assumes flat, mutually exclusive categories. \textit{DocLayNet}~\citep{10.1145/3534678.3539043} and \textit{PubLayNet}~\citep{Zhong2019PubLayNetLD} focus on layout understanding but remain confined to single domains with coarse-grained labels. 
\textit{NYT Corpus}~\citep{ACQUIRED} is text-only and restricted to the news domain. \textit{Web of Science}~\citep{JahaniRad2024HierarchicalTC} focuses on text-centric scientific publications. Therefore, there is an urgent need for a benchmark that reflects the complexity of real business documents, supporting \textbf{hierarchical} classification of \textbf{multi-modal} documents \textbf{across domains}.

To meet this need, we construct \textbf{MMM-Bench}, the first \underline{M}ulti-grained, \underline{M}ulti-domain, \underline{M}ulti-modal document classification \underline{Bench}mark, developed at Alibaba to capture real-world industrial complexity. \textbf{MMM-Bench} offers four key components:\\
$\bullet$ \textbf{Hierarchical taxonomy.} The taxonomy consists of five levels: L1 (Business Function) captures the document's primary purpose; L2 (Business Object) identifies the transacted entity; L3 (Business Stage) encodes its temporal position within a workflow; L4 (Business Form) specifies its structural template; and L5 (Document Class) defines its technical typology. This progressive refinement enables fine-grained evaluation while maintaining semantic coherence across the hierarchy. \\
$\bullet$ \textbf{Multi-modal corpus.} We collect  $5,990$ real-world  documents from $12$ domains in Alibaba. Each document contains rich multi-modal information including scanned images, tables, and text. Every document is manually annotated by domain experts with a complete hierarchical path, enabling systematic evaluation of models on hierarchical classification and fine-grained discrimination.\\
$\bullet$ \textbf{Comprehensive model toolkit}. We integrate 10+ state-of-the-art models into a unified toolkit with one-command installation and deployment, spanning from open-weight large models to API-based large models. This enables researchers to benchmark across diverse architectures without tedious environment setup or code adaptation.\\
$\bullet$ \textbf{Key challenge\&diagnostic insights.} Systematic evaluation unveils four fundamental challenges that define the frontier of this task: \textit{Semantic discrimination difficulty for fine-grained categories},  \textit{domain-sensitive performance fragility}, \textit{cross-modal utilization failure}, and \textit{sample distribution imbalance}. For these challenges, we provide technical and targeted insights.

Our key contributions are summarized as follows:\\
$\bullet$ We construct a five-level document taxonomy to capture authentic business logic, departing fundamentally from the flat, single-domain category spaces of existing benchmarks.\\
$\bullet$ We curate $5,990$ real-world multi-modal documents with expert-annotated hierarchical paths, capturing genuine industrial document distributions.\\
$\bullet$ We integrate 10+ state-of-the-art models (open-weight and API-based) into a unified toolkit with one-command installation and deployment.\\
$\bullet$ We uncover four key challenges by systematic experiments and offer research insights, providing researchers with clear problem formulations and actionable directions for future work.

\section{Related Work}
\subsection{Document Classification Benchmarks}
We categorize existing document classification benchmarks into three types based on their label structure and domain characteristics: flat-label, layout-aware, and hierarchical benchmarks.

\textbf{Flat-label Benchmarks.} Early document classification benchmarks use flat, mutually exclusive category structures that treat each document as belonging to a single class without hierarchical dependencies. For example, \textit{RVL-CDIP} ~\citep{NEURIPS2022_4c0986bd} remains one of the most widely used benchmarks, containing $400,000$ document images across $16$ flat categories. While its scale and diversity have made it a standard for evaluating document image classification, its flat label space fails to capture the nested hierarchical categorization inherent in real-world document management. \textit{Tobacco3482} ~\citep{Kumar2014StructuralSF} provides $3,482$ document images from the tobacco industry, annotated with $10$ flat categories. Despite its real-world origins, its limited scale and domain specificity restrict its utility for general-purpose document classification research. \textit{DocumentNet} ~\citep{DBLP:conf/emnlp/YuMSCHD023} aggregates multiple document datasets into a unified benchmark with $20$ flat categories, but inherits the flat-label limitations of its constituent datasets.

\textbf{Layout-aware Benchmarks.}
Recognizing the importance of visual layout in document understanding, several benchmarks introduce pixel-level layout annotations.
For example,\textit{DocLayNet} ~\citep{10.1145/3534678.3539043} provides $80,000$ manually annotated document pages with bounding box annotations for $11$ layout elements. However, its focus is on layout segmentation rather than document classification, and its documents are drawn primarily from public sources (patents, scientific papers, manuals) with limited domain diversity. \textit{PubLayNet} ~\citep{Zhong2019PubLayNetLD} offers $360,000$ document pages from PubMed Central with layout annotations for 5 element types. While its scale is impressive, it is confined to scientific publications, limiting its applicability to business document classification. \textit{PRImA Layout dataset} ~\citep{5277696} and \textit{TableBank} ~\citep{li-etal-2020-tablebank} focus on layout analysis tasks rather than hierarchical classification, and operate within narrow domains (historical documents and academic tables, respectively).

\textbf{Hierarchical Benchmarks.} Hierarchical document classification benchmarks remain scarce, particularly those combining hierarchy with multi-modal content. For example, \textit{WikiSection} ~\citep{arnold2019sector} provides Wikipedia articles with a two-level hierarchy (31 top-level categories, 650 fine-grained sections). However, it is text-only and focuses on article sections rather than complete documents, limiting its relevance to document classification tasks. \textit{NYT Corpus} ~\citep{ACQUIRED} offers news articles with a hierarchical topic taxonomy, but is text-only and restricted to the news domain. \textit{LEDGAR} ~\citep{tuggener-etal-2020-ledgar} contains $90,000$ legal contract clauses annotated with a two-level hierarchy of $100$ categories. While it demonstrates the value of hierarchical classification in a specialized domain, its focus on clause-level rather than document-level classification, combined with its narrow legal domain, limits broader applicability. \textit{Web of Science} ~\citep{JahaniRad2024HierarchicalTC} provides a hierarchical classification dataset for scientific publications, but like \textit{WikiSection} and \textit{NYT}, it is a text-only and domain-specific dataset.

\begin{table*}[t]
\centering
\caption{Comparison of MMM-Bench with existing document classification benchmarks.}
\label{tab:comparison}
\resizebox{0.9\columnwidth}{!}{
\begin{tabular}{p{2.1cm}p{0.5cm}p{0.9cm}p{1.9cm}ccc}
\toprule
\textbf{Benchmark} & \textbf{Scale} & \textbf{Levels} & \textbf{Multi-modal} & \textbf{Multi-domain} & \textbf{Primary Task} \\
\midrule
\textit{RVL-CDIP} & 400k & 1 (flat) & \checkmark (image) & \ding{55}  & Image classification \\
\textit{Tobacco3482} & 3.5k & 1 (flat) & \checkmark (image) & \ding{55}  & Image classification \\
\textit{DocLayNet} & 80k & 1 (flat) & \checkmark (layout) & \ding{55} & Layout segmentation \\
\textit{PubLayNet} & 360k & 1 (flat) & \checkmark (layout) & \ding{55}  & Layout segmentation \\
\textit{WikiSection}& 100k & 2 & \ding{55} (text-only) & \ding{55} &  Article section classification \\
\textit{NYT Corpus}& 1.8M & 4 & \ding{55} (text-only) & \ding{55} &  News topic classification \\
\textit{LEDGAR} & 90k & 2 & \ding{55} (text-only) & \ding{55} &  Clause classification \\
\textit{Web of Science} & 46k & 3 & \ding{55} (text-only) & \ding{55} &  Publication classification \\
\midrule
\textbf{MMM-Bench} & \textbf{5.9k} & \textbf{5} & \textbf{\checkmark (text+layout+vision)} & \textbf{\checkmark}  & \textbf{Hierarchical classification} \\
\bottomrule
\end{tabular}
}
\vspace{-5mm}
\end{table*}

\subsection{Summary and Positioning}
Table~\ref{tab:comparison} summarizes the key characteristics of existing document classification benchmarks. \textbf{Flat-label benchmarks} assume flat, mutually exclusive categories that fail to capture real-world document hierarchies.
\textbf{Layout-aware benchmarks} provide rich layout annotations and are designed for layout analysis rather than hierarchical classification.
\textbf{Hierarchical benchmarks} offer hierarchical structures but are confined to specific domains (news, legal, scientific) and lack multi-modal content. To the best of our knowledge, no benchmark simultaneously provides: (1) a deeply hierarchical taxonomy, (2) multi-modal content spanning text, layout, and visual information, and (3) multi-domain coverage.

MMM-Bench is the first benchmark to encompass these three critical dimensions, thereby capturing the inherent complexity of real-world document classification. 
First, it provides a deeply hierarchical taxonomy with five levels, capturing the authentic granularity of business document organization. Second, it offers multi-modal richness, with documents containing text, layout structures, visual elements, and tables, enabling systematic evaluation of multi-modal models. Third, it ensures cross-domain coverage, spanning $12$ domains at Alibaba to capture real-world document diversity.

\section{MMM-Bench}
In this section, we introduce \textbf{MMM-Bench} by detailing its multi-level hierarchical taxonomy, multi-modal corpus, and rigorous annotation and quality control procedures.
\subsection{Hierarchical Taxonomy}
The taxonomy is based on industry standards and refined with Alibaba's content management team to reflect real-world business operations.\\
$\bullet$\textbf{Level 1: Business Function.} It partitions the document space according to the primary business function to which a document belongs. It answers the question: ``Which business function does this document serve?'' Examples include Transaction Voucher, Invoice, Contract, Certificate, and Report. It aims to navigate the enterprise document landscape from a macro‑business perspective.\\
$\bullet$\textbf{Level 2: Business Object.} It refines the classification by identifying the core business object that the document records,  answering the question: ``What type of object does this document capture?'' For instance, under the Transaction Voucher function, documents may be categorized by objects such as Goods, Payment, or Travel. It distinguishes documents based on the nature of the transacted entity.\\
$\bullet$\textbf{Level 3: Business Stage.} It specifies the business stage at which the document is generated or used within a transaction lifecycle. It answers the question: ``At which stage of the business process does this document appear?'' Distinctions are made between initiating documents (e.g., Order) and confirming documents (e.g., Receipt/Payment Certificate).
\\
$\bullet$\textbf{Level 4: Business Form.} It introduces the concrete business form or template that a document follows, and answers the question: ``Which specific business form template does this document instantiate?'' Examples include Purchase Order, E‑commerce Order, and Bank Transaction Voucher. It bridges the gap between abstract process roles and tangible document artifacts, grouping documents that share a common structural schema.\\
$\bullet$\textbf{Level 5: Document Class.} It defines the technical classification of a document, and answers the question: ``To which technical document class does this document belong?'' Examples under Bank Transaction Voucher include Bank Receipt and Bank Transfer Record.

\begin{figure}[t]
\centering
\includegraphics[width=0.85\textwidth]{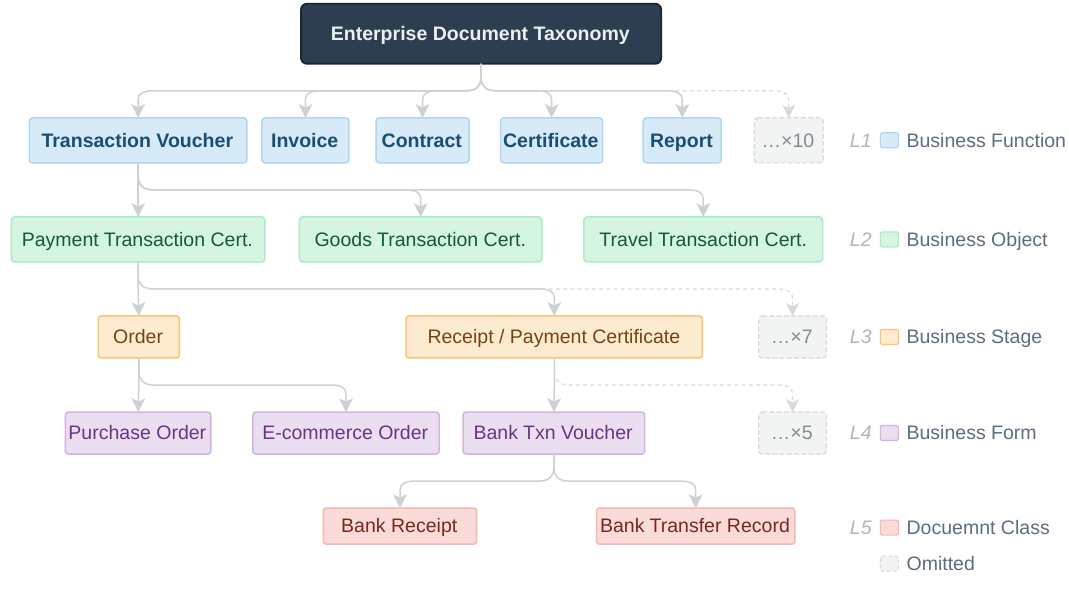}
\vspace{-3mm}
\caption{An example of the hierarchical taxonomy.}
\label{fig:doc_taxonomy}
\vspace{-3mm}
\end{figure}

\begin{table*}[htpb!]
\centering
\caption{Statistics of MMM-Bench hierarchical taxonomy.}
\label{tab:taxonomy_stats}
\resizebox{0.75\columnwidth}{!}{
\begin{tabular}{lcccccr}
\toprule
\textbf{Domain} & \textbf{$L_1$} & \textbf{$L_2$}& \textbf{$L_3$} & \textbf{$L_4$} & \textbf{$L_5$} & Samples\\
\midrule	
FIN (Finance and Accounting) & 10 & 18 & 11 & 2 & 0 & 578 \\
TAX (Taxation) & 11 & 26 & 20 & 4 & 0 & 839 \\
INVEST (Investment and Financing) & 11 & 25 & 21 & 16 & 4 & 1168 \\
LEGAL (Legal and Litigation) & 10 & 22 & 37 & 10 & 3 & 1107 \\
HR (Human Resources) & 10 & 17 & 42 & 2 & 0 & 957 \\
PROCURE (Procurement and Bidding) & 11 & 27 & 54 & 4 & 0 & 1545 \\
CORP (Corporate Governance) & 10 & 19 & 22 & 2 & 0 & 744 \\
REALESTATE (Real Estate and Leasing) & 10 & 20 & 35 & 2 & 0 & 1018 \\
TECH (Technology and Intellectual Property) & 11 & 14 & 28 & 3 & 0 & 939 \\
CONSTRUCT (Construction and Engineering) & 10 & 20 & 15 & 2 & 0 & 705 \\
LOGISTICS (Logistics and Transportation)& 11 & 19 & 11 & 2 & 0 & 612 \\
MARKETING (Marketing and Media)& 10 & 20 & 27 & 2 & 0 & 807 \\
\bottomrule
\end{tabular}
}
\end{table*}

The taxonomy follows a strict tree structure where each node has a single parent, while leaf nodes can be located at L2, L3, L4, or L5 levels. An example of the taxonomy is shown in Figure \ref{fig:doc_taxonomy}. Additionally, a document may belong to multiple domains, because a single business transaction often spans multiple domains. Table \ref{tab:taxonomy_stats} presents the statistics of the hierarchical taxonomy.

\subsection{Multi-modal Document Collection}

We curate $5,990$ real-world multi-modal documents from internal and anonymized repositories within Alibaba, ensuring a high degree of industrial authenticity. The collection process is guided by two principles: \textbf{multi-domain diversity} and \textbf{multi-modal richness}.\\
$\bullet$ \textbf{Multi-domain diversity.} Documents span $12$ distinct domains: FIN, TAX, INVEST, LEGAL, HR, PROCURE, CORP, REALESTATE, TECH, CONSTRUCT, LOGISTICS and  MARKETING.\\
$\bullet$ \textbf{Multi-modal richness.} Each document inherently contains a mixture of modalities: (1) \textbf{natural language text}, including both structured and unstructured content; (2) \textbf{tabular data}, with varying degrees of complexity and formatting; (3) \textbf{visual elements}, such as logos, signatures, and photographs; and (4) \textbf{complex Layouts}, ranging from single-column narratives to multi-column forms with nested tables. The details of the multi-modal richness are reported in Table \ref{tab:multimodal_stats}, and the other statistics of the collection are shown in Table \ref{tab:corpus_stats}.

\begin{table*}[t]
\centering
\begin{minipage}[t]{0.48\textwidth}
\centering
\caption{Multi-modal characteristics of MMM-Bench document collection.We report four categories of modality features.}
\label{tab:multimodal_stats}
\resizebox{\columnwidth}{!}{
\begin{tabular}{lcc}
\toprule
\textbf{Feature} & \textbf{Count} & \textbf{Percentage} \\
\midrule
\textbf{Text content} & & \\
\hspace{0.5cm} Documents with text & 5,977 & 99.78\% \\
\hspace{0.5cm} Avg. text length (tokens) & 450 & — \\
\midrule
\textbf{Tabular content} & & \\
\hspace{0.5cm} Documents containing tables & 2,924 & 48.9\% \\
\hspace{0.5cm} Avg. tables per document & 0.76 & — \\
\midrule
\textbf{Visual elements} & & \\
\hspace{0.5cm} Documents with logos & 453 & 7.6\% \\
\hspace{0.5cm} Documents with stamps/seals & 1,677 & 28.1\% \\
\hspace{0.5cm} Documents with signatures & 370 & 6.2\% \\
\hspace{0.5cm} Documents with color images & 410 & 6.9\% \\
\midrule
\textbf{Layout complexity} & & \\
\hspace{0.5cm} Single-column layout & 5,806 & 96.9\% \\
\hspace{0.5cm} Multi-column layout & 38 & 0.6\% \\
\hspace{0.5cm} Mixed/Complex layout & 131 & 2.2\% \\
\bottomrule
\end{tabular}
}
\end{minipage}
\hfill
\begin{minipage}[t]{0.48\textwidth}
\centering
\caption{Basic statistics of  MMM-Bench.}
\label{tab:corpus_stats}
\resizebox{\columnwidth}{!}{
\begin{tabular}{lcc}
\toprule
\textbf{Statistic} & \textbf{Value} \\
\midrule
Total documents & 5,990 \\
\midrule
\textbf{Visual Tokens} & \\
\hspace{0.5cm} Average visual tokens per document & 1,232 \\
\hspace{0.5cm} Median visual tokens per document & 1,247 \\
\hspace{0.5cm} Maximum visual tokens per document & 1,280 \\
\midrule
\textbf{Text Tokens} & \\
\hspace{0.5cm} Average text tokens per document & 450 \\
\hspace{0.5cm} Median text tokens per document & 430 \\
\hspace{0.5cm} Maximum text tokens per document & 2,215 \\
\midrule
\textbf{Area Px} & \\
\hspace{0.5cm} Average area px per document & 1,646,553 \\
\hspace{0.5cm} Median area px per document & 1,416,000 \\
\hspace{0.5cm} Maximum area px per document & 8,133,600 \\
\midrule
\textbf{Document language} & \\
\hspace{0.5cm} English & 60 (1.0\%) \\
\hspace{0.5cm} Chinese & 5,711 (95.3\%) \\
\hspace{0.5cm} Mixed & 158 (2.6\%) \\
\hspace{0.5cm} Spanish & 25 (0.4\%) \\
\hspace{0.5cm} Portuguese & 20 (0.3\%) \\
\bottomrule
\end{tabular}
}
\end{minipage}
\vspace{-3mm}
\end{table*}

\subsection{Annotation and Quality Control}
Each document is assigned a complete hierarchical path. Given the complexity of the taxonomy and the multi-modal nature of the data, we employ a team of five domain experts. Inspired by the work  ~\citep{Tang2024PDFChatAnnotatorAH,DBLP:conf/cvpr/DengDSLL009},  the annotation workflow consists of three main stages:\\
\textbf{1. Pre-annotation and guideline development.} Before full-scale annotation, a subset of $500$ documents is used to develop a comprehensive annotation guideline which defines the boundaries of each node. Pre-annotation is performed by two senior experts who also act as arbiters in later stages.\\
\textbf{2. Expert annotation.} Each document is independently annotated by three domain experts. Annotators are provided with the complete document information, and are tasked with selecting the correct path through the five-level taxonomy.\\
\textbf{3. Quality control and adjudication:} After the initial annotation, we conduct a rigorous quality control process. Annotations from three experts are compared. For documents where the three annotations agreed, the label is accepted. For the documents where there is a disagreement, the two senior experts, who had led the pre-annotation phase, act as an adjudicator to resolve the conflict.

\textbf{Data Splits.} The dataset of $5,990$ documents is randomly split into training (70\%), validation (10\%), and test (20\%) sets, ensuring a stratified split to preserve the hierarchical and domain-specific distribution across all subsets. This ensures a fair evaluation and prevents data leakage.
\vspace{-2mm}
\section{Experiments}
\label{sec:experiments}
\vspace{-2mm}
\subsection{Experimental Setup}
\label{subsec:setup}
\textbf{Task definition.} Given a  multi-modal document \textit{d}, models are required to predict its labels in the five-level hierarchical taxonomy. We evaluate performance at each level to measure granularity-aware classifications. Additionally, we assess models' generalization ability across different domains.

\textbf{Evaluation metrics.} We use standard classification metrics: \textit{Accuracy} and macro-averaged \textit{F1-score} (macro-F1)~\citep{SOKOLOVA2009427} at each level of the hierarchy. For the overall evaluation, we compute a \textit{Hierarchical F1-score} (HF1)~\citep{8260658} that accounts for the correctness of the entire predicted path from the root to the leaf.

\textbf{Baseline models.} Our evaluation focuses on general-purpose large models to assess their zero-shot and fine-tuned capabilities on MMM-Bench. Accordingly, we select two architectural paradigms:\\
$\bullet$ \textbf{Open-weight models.} We choose several representative open-weight models, e.g., Qwen2.5-VL-7B ~\citep{qwen2025qwen25technicalreport}, Qwen3-VL-8B~\citep{yang2025qwen3technicalreport}, Qwen-3.5-9B/27B/35B-A3B .\\
$\bullet$ \textbf{API-based models.} We also evaluate state-of-the-art commercial API models, e.g., CLAUDE-OPUS (v4.5, v4.6)~\citep{waples2026opus}, GPT-5 (v5.2, v5.4) ~\citep{DBLP:journals/is/Leon26}, Qwen-3-VL235B-A22B, Qwen-3.6-plus, Gemini-3.1-Pro-Preview~\citep{google_gemini}, MiniMax-M2.7~\citep{minimax2026m27}, GLM-5~\citep{glm5team2026glm5vibecodingagentic}, and Kimi-K2.5~\citep{yong2026independentsafetyevaluationkimi}.

\textbf{Implementation details.}
We leverage unified inference settings within each model family. For API models, we set $temperature = 0.1$ and $top\_p = 0.8$, while other parameters (e.g., $top\_k$, $max\_new\_tokens$, and image resolution) are left as default. For open‑weight models, we use the following settings across all checkpoints: $image\_resolution = 262,144$, $cutoff\_len = 20,480$, $max\_new\_tokens = 256$, $temperature = 0.1$, $top\_p = 0.8$, $top\_k = 50$, $num\_beams = 1$, and $repetition\_penalty = 1.0$. Other details are shown in the Section of  Appendices. The difference between models is statistically significant (paired t-test, p<0.05)~\citep{PMID:28904516}, as indicated by $\dagger$.

\vspace{-2mm}
\subsection{Experimental Results and Analysis}
\label{subsec:results}
Based on our proposed benchmark and a diverse set of baseline models, we conduct extensive empirical analyses that reveal four critical challenges. For each challenge, we provide a concrete insight and propose viable technical pathways to guide future research.

\begin{table*}[t]
\centering
\caption{Overall classification performance on MMM-Bench. Significance  is tested on the best-performing model of each metric and $\dagger$ denotes $p<0.05$.}
\label{tab:overall_results}
\resizebox{0.93\columnwidth}{!}{
\begin{tabular}{p{3.5cm}p{0.6cm}p{0.6cm}p{0.6cm}p{0.6cm}p{0.6cm}p{0.6cm}p{0.6cm}p{0.6cm}p{0.6cm}p{0.6cm}p{0.6cm}}
\toprule
\multirow{2}{*}{Model} & \multicolumn{5}{c}{Accuracy @ Level (\%)} & \multicolumn{5}{c}{Macro-F1 @ Level (\%)} & \multirow{2}{*}{HF1} \\
\cmidrule(lr){2-6} \cmidrule(lr){7-11}
                       & L1  & L2 & L3  & L4 & L5 & L1 & L2 & L3  & L4 & L5 & \\
\midrule
\multicolumn{12}{l}{\textit{Weights Models}} \\
\cmidrule(lr){1-12}
Qwen2.5-VL-7B & 65.78$^\dagger$ & 23.25$^\dagger$ & 5.61$^\dagger$ & 0.00$^\dagger$ & 0.00$^\dagger$ & 47.03$^\dagger$ & 18.75$^\dagger$ & 6.39$^\dagger$ & 0.00$^\dagger$ & 0.00$^\dagger$ & 6.66$^\dagger$ \\
Qwen3-VL-8B & 68.78$^\dagger$ & 30.53$^\dagger$ & 11.90$^\dagger$ & 12.75$^\dagger$ & 0.00$^\dagger$ & 69.87$^\dagger$ & 37.63$^\dagger$ & 11.26$^\dagger$ & 14.51$^\dagger$ & 0.00$^\dagger$ & 14.40$^\dagger$ \\
Qwen-3.5-9B & 89.11$^\dagger$ & 66.32$^\dagger$ & 41.75$^\dagger$ & 39.22$^\dagger$ & 13.33$^\dagger$ & 79.80$^\dagger$ & 67.09$^\dagger$ & 41.02$^\dagger$ & 38.82$^\dagger$ & 20.00$^\dagger$ & 38.69$^\dagger$ \\
Qwen-3.5-27B & \underline{93.40} & \underline{77.19} & \underline{57.91} & \underline{57.84} & \underline{53.33} & \underline{82.18} & \underline{76.10} & \underline{56.88} & \underline{57.74} & 50.00 & \underline{51.75} \\
Qwen-3.5-35B-A3B & 91.68$^\dagger$ & 72.19$^\dagger$ & 50.62$^\dagger$ & 49.02$^\dagger$ & \underline{53.33} & 80.33$^\dagger$ & 70.40$^\dagger$ & 49.53$^\dagger$ & 51.89 & \underline{58.93} & 45.38$^\dagger$ \\
\midrule
\addlinespace
\multicolumn{12}{l}{\textit{API Models}} \\
\cmidrule(lr){1-12}
Claude-Opus-4.5 & 95.63 & 82.98 & 71.04$^\dagger$ & 77.45 & 80.00 & 84.39 & 79.67 & 69.74$^\dagger$ & 77.13 & 75.00 & 61.59$^\dagger$ \\
Claude-Opus-4.6 & 95.11 & \underline{83.77} & 74.07$^\dagger$ & 71.57$^\dagger$ & 86.67 & 84.07 & \underline{81.89} & 73.23$^\dagger$ & 71.40$^\dagger$ & 87.50 & 64.01 \\
GPT-5.2 & 94.00$^\dagger$ & 78.07$^\dagger$ & 62.18$^\dagger$ & 69.61$^\dagger$ & 93.33 & 83.27 & 75.97$^\dagger$ & 60.59$^\dagger$ & 68.16$^\dagger$ & \underline{95.00} & 54.35$^\dagger$ \\
GPT-5.4 & 94.60$^\dagger$ & 80.18$^\dagger$ & 64.42$^\dagger$ & 75.49 & 86.67 & 85.98 & 78.11 & 62.00$^\dagger$ & 75.10$^\dagger$ & 87.50 & 55.85$^\dagger$ \\
Qwen-3-VL235B-A22B & 91.51$^\dagger$ & 71.14$^\dagger$ & 52.19$^\dagger$ & 57.84$^\dagger$ & 20.00$^\dagger$ & 85.00 & 71.83$^\dagger$ & 51.99$^\dagger$ & 57.53$^\dagger$ & 21.43$^\dagger$ & 48.13$^\dagger$ \\
Qwen-3.6-plus & 89.02$^\dagger$ & 76.84$^\dagger$ & 63.08$^\dagger$ & 60.78$^\dagger$ & 26.67$^\dagger$ & 82.32$^\dagger$ & 77.38$^\dagger$ & 64.51$^\dagger$ & 63.96$^\dagger$ & 36.67$^\dagger$ & 58.26$^\dagger$ \\
Gemini-3.1-Pro-Preview & \underline{95.80} & 83.51 & \underline{77.22} & \underline{80.39} & \underline{80.00} & \underline{86.10} & 79.94 & \underline{75.91} & \underline{80.14} & 75.00 & \underline{64.98} \\
MiniMax-M2.7 & 92.37$^\dagger$ & 76.4$^\dagger$ & 56.23$^\dagger$ & 49.02$^\dagger$ & 20.00$^\dagger$ & 81.30$^\dagger$ & 76.14$^\dagger$ & 58.54$^\dagger$ & 48.62$^\dagger$ & 26.67$^\dagger$ & 51.36$^\dagger$ \\
GLM-5 & 73.24$^\dagger$ & 67.19$^\dagger$ & 60.72$^\dagger$ & 49.02$^\dagger$ & 13.33$^\dagger$ & 77.81$^\dagger$ & 74.23$^\dagger$ & 63.95$^\dagger$ & 56.23$^\dagger$ & 20.00$^\dagger$ & 54.67$^\dagger$ \\
Kimi-K2.5 & 94.08$^\dagger$ & 79.21$^\dagger$ & 65.32$^\dagger$ & 53.92$^\dagger$ & 46.67 & 84.04 & 76.62$^\dagger$ & 64.59$^\dagger$ & 53.27$^\dagger$ & 51.67$^\dagger$ & 57.06$^\dagger$ \\
\bottomrule
\end{tabular}
}
\vspace{-3mm}
\end{table*}

\vspace{-2mm}
\subsubsection{Challenge One: Semantic Discrimination Difficulty for Fine-Grained Categories}
Table~\ref{tab:overall_results} presents the overall experimental results. It can be seen that models perform better at L1 and L2 than at L3, L4 and L5. This is because the higher-level categories (L1: Business Function, L2: Business Object) correspond to broad, well‑differentiated concepts that are relatively easy to recognize from documents. In contrast, L3 (Business Stage), L4 (Business Form), and L5 (Document Class) involve finer distinctions. For example, at L3, models must differentiate between an initiating document (e.g., a Purchase Order) and a confirming document (e.g., a Payment Receipt). But models often struggle to demarcate semantic boundaries between fine-grained categories.

\textbf{Insight: joint hierarchical training.} We argue that future models should be trained under a joint hierarchical objective. The core idea is to explicitly co-optimize models for all granularity levels simultaneously,  from L1 down to L5 within a unified architecture. This method aims to forge feature representations that are not merely adequate for any single level, but are holistically discriminative across the entire hierarchy. This approach seeks to transform the representational conflict into a synergistic learning process, thereby mitigating the performance degradation at fine granularity.
\vspace{-2mm}
\subsubsection{Challenge Two: Domain-Sensitive Performance Fragility} 
\begin{figure}[t]
\centering
\begin{subfigure}[t]{0.49\textwidth}
\centering
\includegraphics[width=\textwidth]{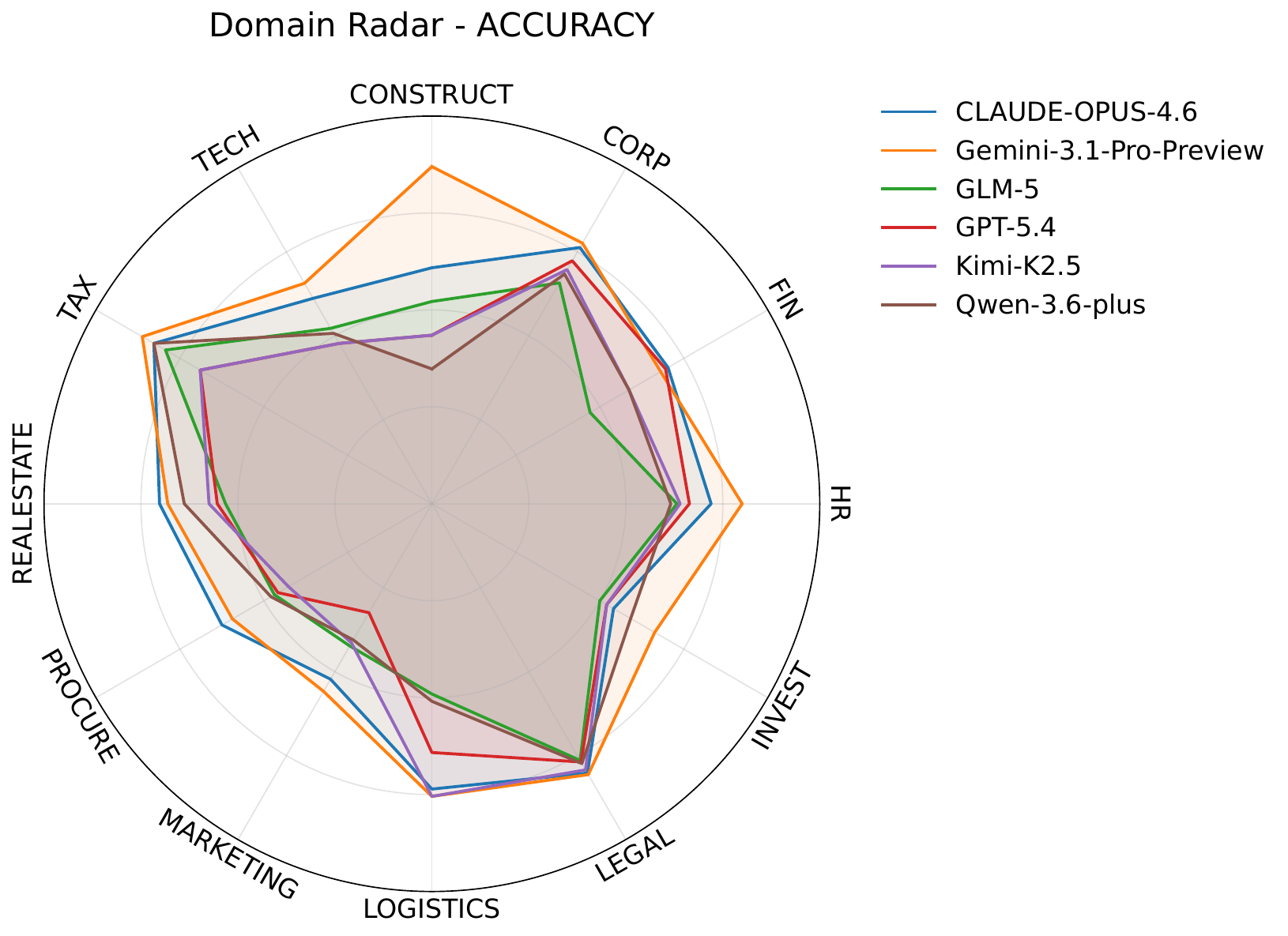}
\caption{Accuracy}
\end{subfigure}
\hfill
\begin{subfigure}[t]{0.49\textwidth}
\centering
\includegraphics[width=\textwidth]{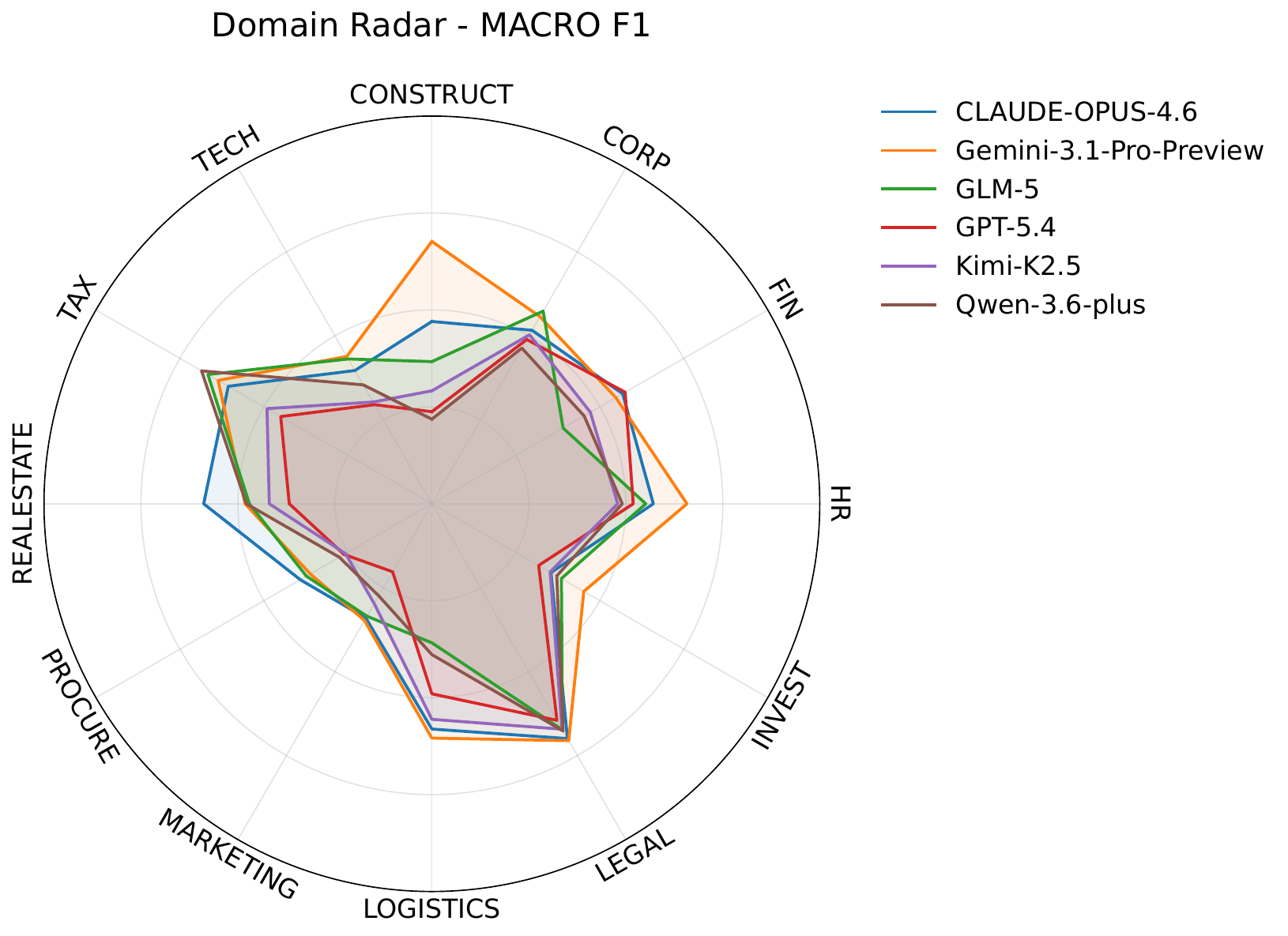}
\caption{Macro-F1}
\end{subfigure}
\caption{The performance of API-based large  models with different domains.}
\label{fig:api_model_domain_radar_all_metrics}
\vspace{-3mm}
\end{figure}

\begin{figure}[t]
\centering
\begin{subfigure}[t]{0.49\textwidth}
\centering
\includegraphics[width=\textwidth]{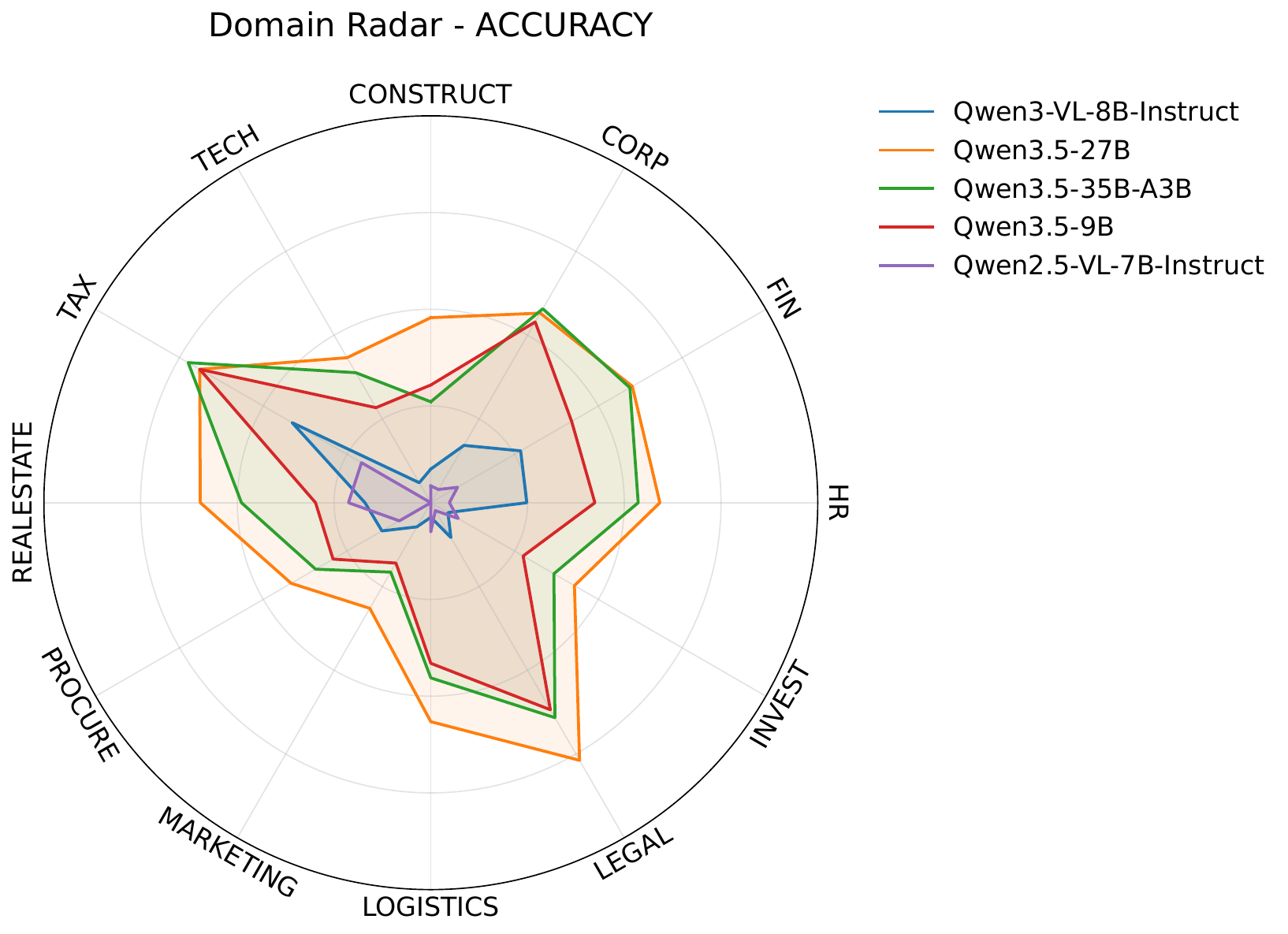}
\caption{Accuracy}
\end{subfigure}
\hfill
\begin{subfigure}[t]{0.49\textwidth}
\centering
\includegraphics[width=\textwidth]{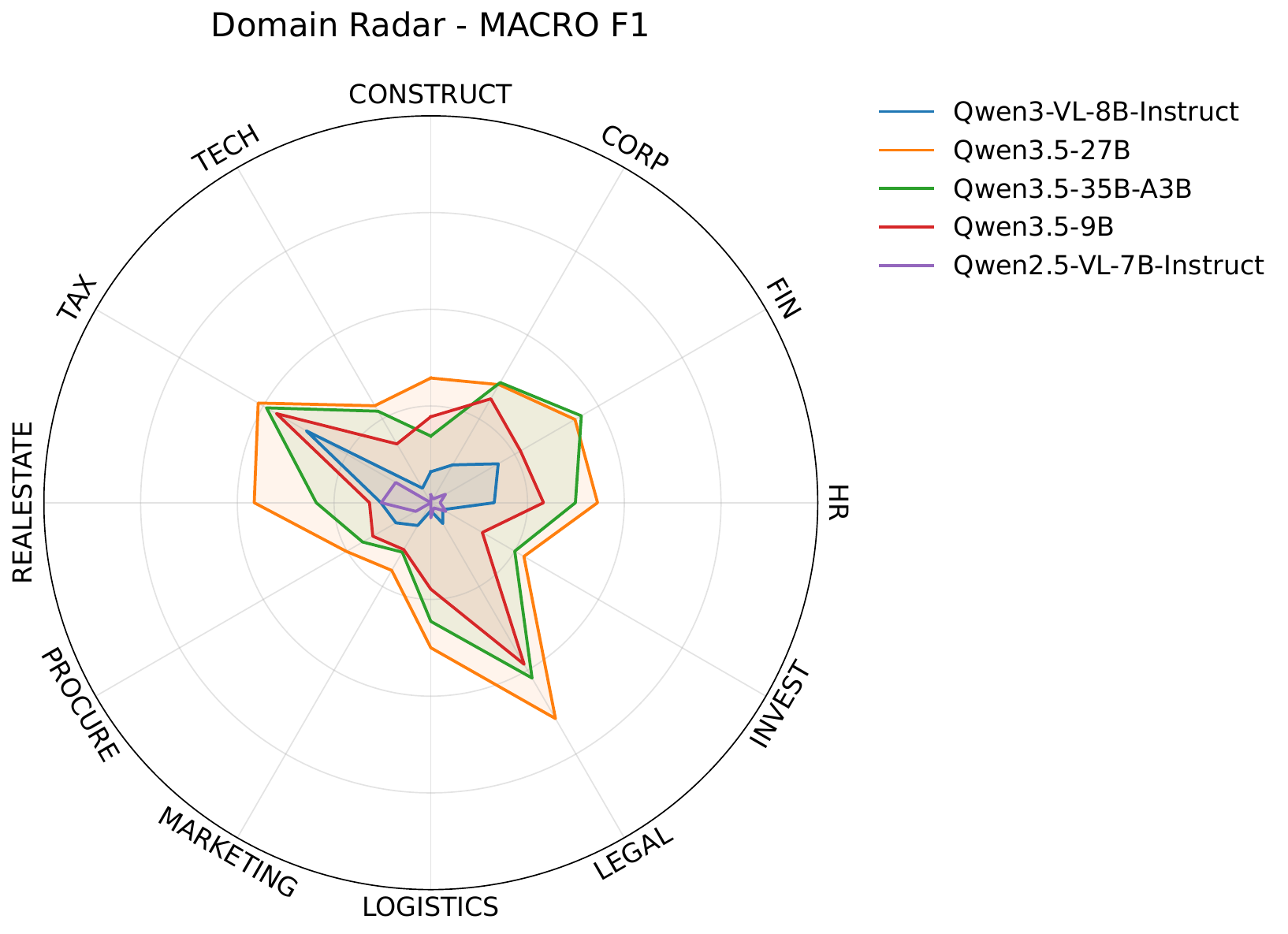}
\caption{Macro-F1}
\end{subfigure}
\caption{The performance of open-weight large models with different domains.}
\label{fig:weights_model_domain_radar_all_metrics}
\vspace{-3mm}
\end{figure}

Figure
\ref{fig:api_model_domain_radar_all_metrics} and \ref{fig:weights_model_domain_radar_all_metrics} show the performance of API-based and open-weight models in different domains. It can be seen that models generally achieve the highest performance in the LEGAL domain, followed by CORP and TAX domains, while struggling more in the CONSTRUCT and MARKETING domains. This trend reveals that model performance is closely tied to domain-specific characteristics. The LEGAL domain benefits from highly structured, text-dense documents with distinctive legal terminology. In contrast, the MARKETING domain, comprising diverse forms, reports, and certificates with highly variable layouts and sparse textual cues, presents a greater challenge.

\textbf{Insight: domain-adaptive modeling.} To address the above challenge, future work can shift from a ``domain-agnostic'' to a ``domain-adaptive'' design paradigm,  enabling models to perceive the intrinsic characteristics of document domains and dynamically adjust their inference mechanisms. Specifically, viable technical pathways include: (1) adopting domain-adversarial training to decouple domain-invariant essential representations from domain-specific stylistic representations; (2) introducing meta-learning or prompt-learning frameworks for rapid few-shot adaptation to new document domains; and (3) designing mixture-of-experts based domain-adaptive architectures, where a router network dynamically activates expert modules to match the input document's domain. 
\vspace{-2mm}
\subsubsection{Challenge Three: Cross-Modal Utilization Failure} We employ supervised fine-tuning (SFT)~\citep{DBLP:conf/emnlp/YeYNLZGHWSF25} on Qwen2.5-VL-7B, Qwen3-VL-8B, and Qwen3.5-9B to examine their capacity for leveraging different modalities, and show  the experimental results in Table \ref{tab:modality_ablation}. In the table, ``$\surd$'' denotes that a given modality is included as the input of models, while ``--'' denotes that it is excluded. We can see  that for the Qwen2.5-VL-7B-SFT, Qwen3-VL-8B-SFT, and Qwen3.5-9B-SFT models, the text-only configuration outperforms all other configurations that include one or more additional modalities. This illustrates that current models, under standard SFT, fail to benefit from additional modalities, i.e., cross-modal utilization failure.

\begin{table*}[t]
\centering
\caption{Modality ablation results (Macro-F1, \%) across hierarchical levels. }
\label{tab:modality_ablation}
\resizebox{0.9\textwidth}{!}{%
\begin{tabular}{p{3.8cm}ccc|cccccc}
\toprule
\textbf{Model} & \textbf{Text} & \textbf{Image} & \textbf{Layout} & \textbf{L1} & \textbf{L2} & \textbf{L3} & \textbf{L4} & \textbf{L5} & \textbf{HF1} \\
\midrule
\multirow{5}{*}{Qwen2.5-VL-7B-SFT}
& $\surd$ & $\surd$ & $\surd$ & 91.27 & 89.05 & 86.69 & 93.58 & \underline{96.43} & 83.98 \\
& $\surd$ & -- & $\surd$ & 84.96 & 87.15 & 85.70 & 80.62 & 87.86 & 81.21 \\
& -- & $\surd$ & -- & 81.43 & 73.78 & 69.06 & 90.37 & 86.67 & 68.50 \\
& $\surd$ & $\surd$ & -- & 89.73 & 93.37 & 92.41 & 95.03 & 95.00 & 88.46 \\
& $\surd$ & -- & -- & \underline{97.36} & \underline{96.27} & \underline{94.85} & \underline{97.52} & 87.50 & \underline{93.20} \\
\midrule

\multirow{5}{*}{Qwen3-VL-8B-SFT}
& $\surd$ & $\surd$ & $\surd$ & 92.17 & 91.18 & 90.16 & 90.76 & \underline{96.43} & 86.80 \\
& $\surd$ & -- & $\surd$ & 85.78 & 88.73 & 88.73 & 81.15 & 83.10 & 84.04 \\
& -- & $\surd$ & -- & 92.05 & 84.71 & 79.27 & 92.18 & 95.00 & 78.52 \\
& $\surd$ & $\surd$ & -- & 92.93 & 95.08 & 95.06 & 97.74 & 95.00 & 92.74 \\
& $\surd$ & -- & -- & \underline{98.44} & \underline{96.39} & \underline{95.66} & \underline{99.17} & 95.00 & \underline{94.17} \\
\midrule

\multirow{5}{*}{Qwen3.5-9B-SFT}
& $\surd$ & $\surd$ & $\surd$ & 91.62 & 92.17 & 89.37 & 93.75 & 96.43 & 86.79 \\
& $\surd$ & -- & $\surd$ & 88.28 & 87.04 & 86.13 & 88.35 & 92.86 & 82.86 \\
& -- & $\surd$ & -- & 89.63 & 86.86 & 81.56 & 90.26 & 80.36 & 79.28 \\
& $\surd$ & $\surd$ & -- & 94.80 & 92.07 & 90.46 & 91.08 & 95.00 & 88.03 \\
& $\surd$ & -- & -- & \underline{95.07} & \underline{94.94} & \underline{94.68} & \underline{96.15} & \underline{100.00} & \underline{92.88} \\
\bottomrule
\end{tabular}%
}
\vspace{-3mm}
\end{table*}

\textbf{Insight: cross-modal attention mechanisms.} For the above challenge,  we argue that future models should move beyond simple modality concatenation or shallow fusion strategies and instead develop more sophisticated cross-modal attention architectures. As a result, models can dynamically weigh the relevance of each modality on a per-instance and per-level basis, and suppress noisy or uninformative modal signals, while amplifying complementary information. Specifically, viable technical pathways include: (1) cross-modal gated attention, where learnable gates control the information flow from each modality before fusion; (2) hierarchy-aware cross-modal attention, which adapts attention patterns according to the granularity level (e.g., relying more on vision at coarse levels and text at fine levels); and (3) iterative cross-modal refinement, where modalities alternately refine each other's representations through multiple interaction cycles.

\subsubsection{Challenge Four: Sample Distribution Imbalance} 
\begin{figure}[t]
\centering
\includegraphics[width=0.8\columnwidth]{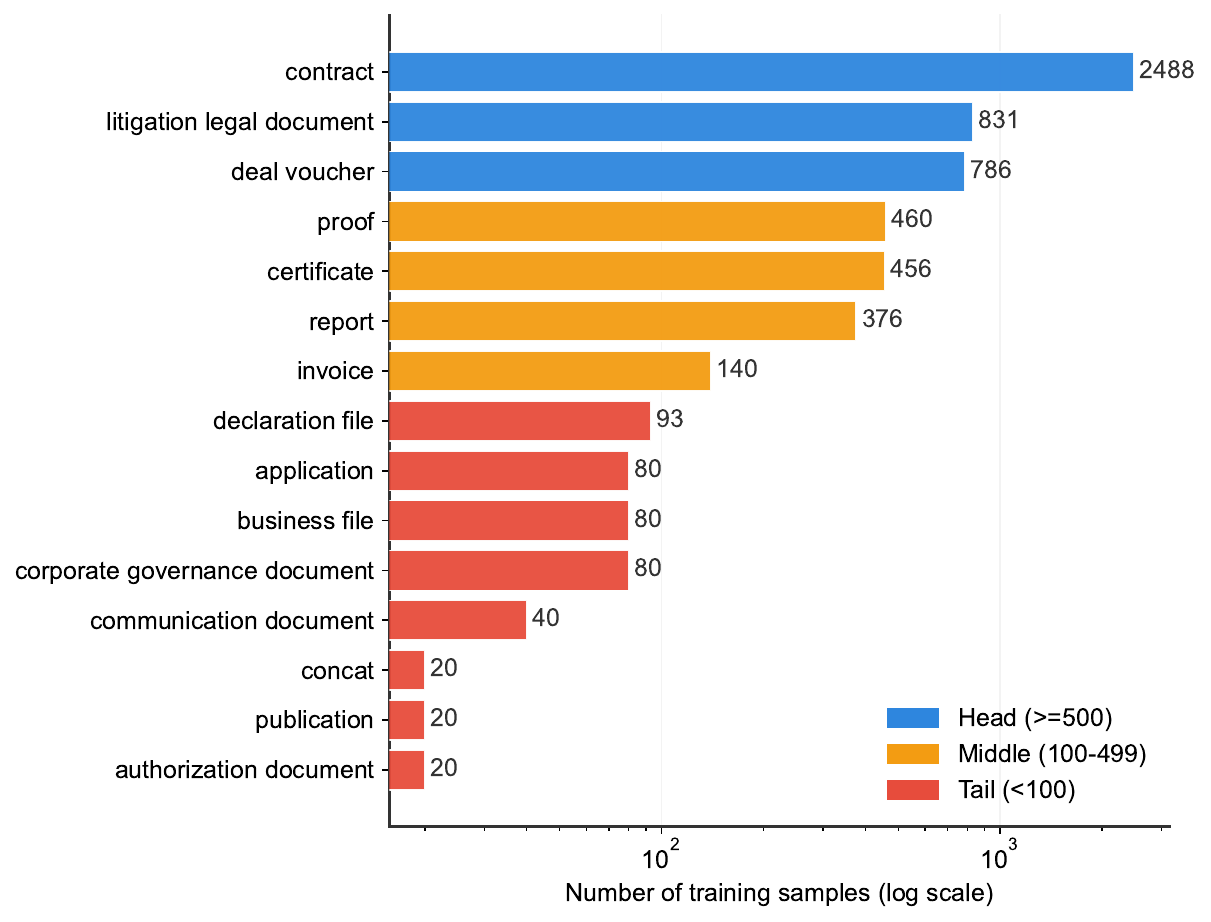}
\vspace{-3mm}
\caption{Long-tailed distribution of training samples at L1 level.}
\label{fig:class_distribution}
\vspace{-3mm}
\end{figure}

We categorize all document instances into three groups based on the number of samples $n$: Head (top $20\%$, $n>=500$), Middle (middle $60\%, 100<=n<500$), and Tail (bottom $20\%, n<100$). The statistical results are shown in Figure~\ref{fig:class_distribution}. It can be seen that of the $15$ L1 categories, $8$ are classified into the Tail group, $4$ into the Middle group, and the remaining $3$ into the Head group, which confirms the long-tailed nature of MMM-Bench. 
Besides, Table \ref{tab:performance_by_frequency} reports the performance of Qwen3.5-9B-SFT across the three groups. The performance in the Tail group is lower than that in the Head and Middle groups. This illustrates that models struggle to cope with the sample distribution imbalance problem.

\begin{table*}[htbp]
\centering
\caption{Performance of Qwen3.5-9B-SFT on Head, Middle, and Tail groups.}
\label{tab:performance_by_frequency}
\resizebox{0.7\columnwidth}{!}{
\begin{tabular}{p{2.8cm}p{2.2cm}p{2.6cm}p{2.6cm}}
\toprule
\textbf{Class Group} & \textbf{Condition} & \textbf{Avg. Samples} & \textbf{Macro-F1} \\
\midrule
Head & $n \geq 500$ & 1,368 & 92.1\% \\
Middle & $100 \leq n < 500$ & 358 & 87.7\% \\
Tail & $n < 100$ & 54 & 60.2\% \\
\bottomrule
\end{tabular}
}
\end{table*}

\textbf{Insight: hierarchical knowledge transfer.} To address this challenge, we argue that future models should leverage hierarchical knowledge transfer, i.e., explicitly exploiting the semantic relationships between coarse and fine-grained categories to mitigate data imbalance. Specifically, viable technical pathways include: (1) hierarchical parameter sharing, where lower-level classifiers share or inherit parameters from higher-level ones, enabling few-shot tail categories to benefit from well-learned coarse-grained representations; (2) hierarchical contrastive learning, which enforces that embeddings of fine-grained categories respect the taxonomic structure, pulling together semantically related classes across levels while separating distinct ones; and (3) hierarchical data augmentation, where synthetic samples for tail categories are generated by perturbing or interpolating coarse-grained representations from parent nodes.

\vspace{-2mm}
\section{Conclusion}
We introduce MMM-Bench, the first benchmark that jointly captures multi-level taxonomy, multi-domain coverage, and multi-modal content for document classification. Unlike existing benchmarks trapped in flat labels or single-domain settings, MMM-Bench features a five-level taxonomy spanning $12$ real-world domains with $5,990$ expert-annotated documents. Systematic evaluation of $10+$ state-of-the-art models reveals four fundamental challenges: semantic discrimination difficulty for fine-grained categories, domain-sensitive performance fragility, cross-modal utilization failure, and sample distribution imbalance. For these challenges, we propose insights: joint hierarchical training, domain-adaptive modeling,  cross-modal attention mechanisms, and hierarchical knowledge transfer. The challenges and insights provide researchers with clear problem formulations and actionable directions for future work. Our current experiments focus exclusively on large language models, and we have not evaluated non-large-model approaches. Evaluating such methods constitutes an important direction for future work.

\bibliographystyle{plainnat}
\bibliography{sample-base}

@String{Computing = "Computing" }

@String{Computer = "{IEEE} Computer" }

@inproceedings{DBLP:conf/emnlp/YeYNLZGHWSF25,
  author       = {Junjie Ye and
                  Yuming Yang and
                  Yang Nan and
                  Shuo Li and
                  Qi Zhang and
                  Tao Gui and
                  Xuanjing Huang and
                  Peng Wang and
                  Zhongchao Shi and
                  Jianping Fan},
  editor       = {Christos Christodoulopoulos and
                  Tanmoy Chakraborty and
                  Carolyn Rose and
                  Violet Peng},
  title        = {Analyzing the Effects of Supervised Fine-Tuning on Model Knowledge
                  from Token and Parameter Levels},
  booktitle    = {Proceedings of the 2025 Conference on Empirical Methods in Natural
                  Language Processing, {EMNLP} 2025, Suzhou, China, November 4-9, 2025},
  pages        = {471--513},
  publisher    = {Association for Computational Linguistics},
  year         = {2025},
  url          = {https://doi.org/10.18653/v1/2025.emnlp-main.25},
  doi          = {10.18653/V1/2025.EMNLP-MAIN.25},
  bibsource    = {dblp computer science bibliography, https://dblp.org}
}

@article{NAKAYAMA2024102717,
title = {Organic transformation of ERP documentation practices: Moving from archival records to dialogue-based, agile throwaway documents},
journal = {International Journal of Information Management},
volume = {74},
pages = {102717},
year = {2024},
issn = {0268-4012},
doi = {https://doi.org/10.1016/j.ijinfomgt.2023.102717},
url = {https://www.sciencedirect.com/science/article/pii/S0268401223000981},
author = {Makoto Nakayama and Eli Hustad and Norma Sutcliffe and Merri Beckfield}
}

@inproceedings{NEURIPS2022_4c0986bd,
 author = {Larson, Stefan and Lim, Yi Yang Gordon and Ai, Yutong and Kuang, David and Leach, Kevin},
 booktitle = {Advances in Neural Information Processing Systems},
 editor = {S. Koyejo and S. Mohamed and A. Agarwal and D. Belgrave and K. Cho and A. Oh},
 pages = {11673--11685},
 publisher = {Curran Associates, Inc.},
 title = {Evaluating Out-of-Distribution Performance on Document Image Classifiers},
 url = {https://proceedings.neurips.cc/paper_files/paper/2022/file/4c0986bd04d747745beba3752bdf4d9d-Paper-Datasets_and_Benchmarks.pdf},
 volume = {35},
 year = {2022}
}

@article{Zhong2019PubLayNetLD,
  title={PubLayNet: Largest Dataset Ever for Document Layout Analysis},
  author={Xu Zhong and Jianbin Tang and Antonio Jimeno-Yepes},
  journal={2019 International Conference on Document Analysis and Recognition (ICDAR)},
  year={2019},
  pages={1015-1022},
  url={https://api.semanticscholar.org/CorpusID:201124789}
}

@inproceedings{10.1145/3534678.3539043,
author = {Pfitzmann, Birgit and Auer, Christoph and Dolfi, Michele and Nassar, Ahmed S. and Staar, Peter},
title = {DocLayNet: A Large Human-Annotated Dataset for Document-Layout Segmentation},
year = {2022},
isbn = {9781450393850},
publisher = {Association for Computing Machinery},
address = {New York, NY, USA},
url = {https://doi.org/10.1145/3534678.3539043},
doi = {10.1145/3534678.3539043},
booktitle = {Proceedings of the 28th ACM SIGKDD Conference on Knowledge Discovery and Data Mining},
pages = {3743–3751},
numpages = {9},
location = {Washington DC, USA},
series = {KDD '22}
}

@article{Kumar2014StructuralSF,
  title={Structural similarity for document image classification and retrieval},
  author={Jayant Kumar and Peng Ye and David S. Doermann},
  journal={Pattern Recognit. Lett.},
  year={2014},
  volume={43},
  pages={119-126},
  url={https://api.semanticscholar.org/CorpusID:207329118}
}

@inproceedings{DBLP:conf/emnlp/YuMSCHD023,
  author       = {Lijun Yu and
                  Jin Miao and
                  Xiaoyu Sun and
                  Jiayi Chen and
                  Alexander G. Hauptmann and
                  Hanjun Dai and
                  Wei Wei},
  editor       = {Mingxuan Wang and
                  Imed Zitouni},
  title        = {DocumentNet: Bridging the Data Gap in Document Pre-training},
  booktitle    = {Proceedings of the 2023 Conference on Empirical Methods in Natural
                  Language Processing: {EMNLP} 2023 - Industry Track, Singapore, December
                  6-10, 2023},
  pages        = {707--722},
  publisher    = {Association for Computational Linguistics},
  year         = {2023},
  url          = {https://doi.org/10.18653/v1/2023.emnlp-industry.66},
  doi          = {10.18653/V1/2023.EMNLP-INDUSTRY.66},
  bibsource    = {dblp computer science bibliography, https://dblp.org}
}

@inproceedings{li-etal-2020-tablebank,
    title = "{T}able{B}ank: Table Benchmark for Image-based Table Detection and Recognition",
    author = "Li, Minghao  and
      Cui, Lei  and
      Huang, Shaohan  and
      Wei, Furu  and
      Zhou, Ming  and
      Li, Zhoujun",
    editor = "Calzolari, Nicoletta  and
      B{\'e}chet, Fr{\'e}d{\'e}ric  and
      Blache, Philippe  and
      Choukri, Khalid  and
      Cieri, Christopher  and
      Declerck, Thierry  and
      Goggi, Sara  and
      Isahara, Hitoshi  and
      Maegaard, Bente  and
      Mariani, Joseph  and
      Mazo, H{\'e}l{\`e}ne  and
      Moreno, Asuncion  and
      Odijk, Jan  and
      Piperidis, Stelios",
    booktitle = "Proceedings of the Twelfth Language Resources and Evaluation Conference",
    month = may,
    year = "2020",
    address = "Marseille, France",
    publisher = "European Language Resources Association",
    url = "https://aclanthology.org/2020.lrec-1.236/",
    pages = "1918--1925",
    language = "eng",
    ISBN = "979-10-95546-34-4"
}

@INPROCEEDINGS{5277696,
  author={Antonacopoulos, Apostolos and Bridson, David and Papadopoulos, Christos and Pletschacher, Stefan},
  booktitle={2009 10th International Conference on Document Analysis and Recognition}, 
  title={A Realistic Dataset for Performance Evaluation of Document Layout Analysis}, 
  year={2009},
  volume={},
  number={},
  pages={296-300}
  }

@article{arnold2019sector,
  author = {Arnold, Sebastian and Schneider, Rudolf and Cudré-Mauroux, Philippe and Gers, Felix A. and Löser, Alexander},
  title = {SECTOR: A Neural Model for Coherent Topic Segmentation and Classification},
  journal = {Transactions of the Association for Computational Linguistics},
  volume = {7},
  pages = {169-184},
  year = {2019},
  doi = {10.1162/tacl\_a\_00261}
}

@inproceedings{tuggener-etal-2020-ledgar,
    title = "{LEDGAR}: A Large-Scale Multi-label Corpus for Text Classification of Legal Provisions in Contracts",
    author = {Tuggener, Don  and
      von D{\"a}niken, Pius  and
      Peetz, Thomas  and
      Cieliebak, Mark},
    editor = "Calzolari, Nicoletta  and
      B{\'e}chet, Fr{\'e}d{\'e}ric  and
      Blache, Philippe  and
      Choukri, Khalid  and
      Cieri, Christopher  and
      Declerck, Thierry  and
      Goggi, Sara  and
      Isahara, Hitoshi  and
      Maegaard, Bente  and
      Mariani, Joseph  and
      Mazo, H{\'e}l{\`e}ne  and
      Moreno, Asuncion  and
      Odijk, Jan  and
      Piperidis, Stelios",
    booktitle = "Proceedings of the Twelfth Language Resources and Evaluation Conference",
    month = may,
    year = "2020",
    address = "Marseille, France",
    publisher = "European Language Resources Association",
    url = "https://aclanthology.org/2020.lrec-1.155/",
    pages = "1235--1241",
    language = "eng",
    ISBN = "979-10-95546-34-4"
}

@article{JahaniRad2024HierarchicalTC,
  title={Hierarchical text classification for web of science scientific fields},
  author={Pouyan Jahani Rad and Mahdi Bahaghighat},
  journal={Facta universitatis - series: Electronics and Energetics},
  year={2024},
  url={https://api.semanticscholar.org/CorpusID:276015364}
}

@misc{qwen2025qwen25technicalreport,
      title={Qwen2.5 Technical Report}, 
      author={Qwen and : and An Yang and Baosong Yang and Beichen Zhang and Binyuan Hui and Bo Zheng and Bowen Yu and Chengyuan Li and Dayiheng Liu and Fei Huang and Haoran Wei and Huan Lin and Jian Yang and Jianhong Tu and Jianwei Zhang and Jianxin Yang and Jiaxi Yang and Jingren Zhou and Junyang Lin and Kai Dang and Keming Lu and Keqin Bao and Kexin Yang and Le Yu and Mei Li and Mingfeng Xue and Pei Zhang and Qin Zhu and Rui Men and Runji Lin and Tianhao Li and Tianyi Tang and Tingyu Xia and Xingzhang Ren and Xuancheng Ren and Yang Fan and Yang Su and Yichang Zhang and Yu Wan and Yuqiong Liu and Zeyu Cui and Zhenru Zhang and Zihan Qiu},
      year={2025},
      eprint={2412.15115},
      archivePrefix={arXiv},
      primaryClass={cs.CL},
      url={https://arxiv.org/abs/2412.15115}, 
}

@article{DBLP:journals/is/Leon26,
  author       = {Maikel Le{\'{o}}n},
  title        = {{GPT-5} and open-weight large language models: Advances in reasoning,
                  transparency, and control},
  journal      = {Inf. Syst.},
  volume       = {136},
  pages        = {102620},
  year         = {2026},
  url          = {https://doi.org/10.1016/j.is.2025.102620},
  doi          = {10.1016/J.IS.2025.102620},
  bibsource    = {dblp computer science bibliography, https://dblp.org}
}

@inproceedings{DBLP:conf/sigmod/Sheth95,
  author       = {Amit P. Sheth},
  editor       = {Michael J. Carey and
                  Donovan A. Schneider},
  title        = {Workflow Automation: Applications, Technology, and Research (Tutorial)},
  booktitle    = {Proceedings of the 1995 {ACM} {SIGMOD} International Conference on
                  Management of Data, San Jose, California, USA, May 22-25, 1995},
  pages        = {469},
  publisher    = {{ACM} Press},
  year         = {1995},
  url          = {https://doi.org/10.1145/223784.223882},
  doi          = {10.1145/223784.223882},
  bibsource    = {dblp computer science bibliography, https://dblp.org}
}

@misc{ACQUIRED,
      recid = {103638},
      author = {Sandhaus, Evan},
      title = {The New York Times Annotated Corpus},
      howpublished = {Linguistic Data Consortium},
      year = {2008},
      note = {LDC2008T19}
}

@inproceedings{DBLP:conf/sigir/Zarri84,
  author       = {Gian Piero Zarri},
  editor       = {C. J. van Rijsbergen},
  title        = {Some Remarks About the Inference Techniques of RESEDA, an "Intelligent"
                  Information Retrieval System},
  booktitle    = {Research and Development in Information Retrieval, Proceedings of
                  the Third Joint {BCS/ACM} Symposium on Research and Development in
                  Information Retrieval, Cambridge, UK, 2-6 July 1984},
  pages        = {281--300},
  publisher    = {Cambridge University Press, on behalf of the British Computer Society
                  (copyright by British Informatics Society Ltd, a subsidiary of the
                  British Computer Society)},
  year         = {1984},
  url          = {http://dl.acm.org/citation.cfm?id=636824},
  bibsource    = {dblp computer science bibliography, https://dblp.org}
}

@article{DBLP:journals/tip/GerekCTA99,
  author       = {{\"{O}}mer Nezih Gerek and
                  A. Enis {\c{C}}etin and
                  Ahmed H. Tewfik and
                  Volkan Atalay},
  title        = {Subband domain coding of binary textual images for document archiving},
  journal      = {{IEEE} Trans. Image Process.},
  volume       = {8},
  number       = {10},
  pages        = {1438--1446},
  year         = {1999},
  url          = {https://doi.org/10.1109/83.791969},
  doi          = {10.1109/83.791969},
  bibsource    = {dblp computer science bibliography, https://dblp.org}
}

@article{Tang2024PDFChatAnnotatorAH,
  title={PDFChatAnnotator: A Human-LLM Collaborative Multi-Modal Data Annotation Tool for PDF-Format Catalogs},
  author={Yi Tang and Chia-Ming Chang and Xi Yang},
  journal={Proceedings of the 29th International Conference on Intelligent User Interfaces},
  year={2024},
  url={https://api.semanticscholar.org/CorpusID:268979980}
}

@misc{yang2025qwen3technicalreport,
      title={Qwen3 Technical Report}, 
      author={An Yang and Anfeng Li and Baosong Yang and Beichen Zhang and Binyuan Hui and Bo Zheng and Bowen Yu and Chang Gao and Chengen Huang and Chenxu Lv and Chujie Zheng and Dayiheng Liu and Fan Zhou and Fei Huang and Feng Hu and Hao Ge and Haoran Wei and Huan Lin and Jialong Tang and Jian Yang and Jianhong Tu and Jianwei Zhang and Jianxin Yang and Jiaxi Yang and Jing Zhou and Jingren Zhou and Junyang Lin and Kai Dang and Keqin Bao and Kexin Yang and Le Yu and Lianghao Deng and Mei Li and Mingfeng Xue and Mingze Li and Pei Zhang and Peng Wang and Qin Zhu and Rui Men and Ruize Gao and Shixuan Liu and Shuang Luo and Tianhao Li and Tianyi Tang and Wenbiao Yin and Xingzhang Ren and Xinyu Wang and Xinyu Zhang and Xuancheng Ren and Yang Fan and Yang Su and Yichang Zhang and Yinger Zhang and Yu Wan and Yuqiong Liu and Zekun Wang and Zeyu Cui and Zhenru Zhang and Zhipeng Zhou and Zihan Qiu},
      year={2025},
      eprint={2505.09388},
      archivePrefix={arXiv},
      primaryClass={cs.CL},
      url={https://arxiv.org/abs/2505.09388}, 
}

@misc{glm5team2026glm5vibecodingagentic,
      title={GLM-5: from Vibe Coding to Agentic Engineering}, 
      author={GLM-5-Team and : and Aohan Zeng and Xin Lv and Zhenyu Hou and Zhengxiao Du and Qinkai Zheng and Bin Chen and Da Yin and Chendi Ge and Chenghua Huang and Chengxing Xie and Chenzheng Zhu and Congfeng Yin and Cunxiang Wang and Gengzheng Pan and Hao Zeng and Haoke Zhang and Haoran Wang and Huilong Chen and Jiajie Zhang and Jian Jiao and Jiaqi Guo and Jingsen Wang and Jingzhao Du and Jinzhu Wu and Kedong Wang and Lei Li and Lin Fan and Lucen Zhong and Mingdao Liu and Mingming Zhao and Pengfan Du and Qian Dong and Rui Lu and Shuang-Li and Shulin Cao and Song Liu and Ting Jiang and Xiaodong Chen and Xiaohan Zhang and Xuancheng Huang and Xuezhen Dong and Yabo Xu and Yao Wei and Yifan An and Yilin Niu and Yitong Zhu and Yuanhao Wen and Yukuo Cen and Yushi Bai and Zhongpei Qiao and Zihan Wang and Zikang Wang and Zilin Zhu and Ziqiang Liu and Zixuan Li and Bojie Wang and Bosi Wen and Can Huang and Changpeng Cai and Chao Yu and Chen Li and Chengwei Hu and Chenhui Zhang and Dan Zhang and Daoyan Lin and Dayong Yang and Di Wang and Ding Ai and Erle Zhu and Fangzhou Yi and Feiyu Chen and Guohong Wen and Hailong Sun and Haisha Zhao and Haiyi Hu and Hanchen Zhang and Hanrui Liu and Hanyu Zhang and Hao Peng and Hao Tai and Haobo Zhang and He Liu and Hongwei Wang and Hongxi Yan and Hongyu Ge and Huan Liu and Huanpeng Chu and Jia'ni Zhao and Jiachen Wang and Jiajing Zhao and Jiamin Ren and Jiapeng Wang and Jiaxin Zhang and Jiayi Gui and Jiayue Zhao and Jijie Li and Jing An and Jing Li and Jingwei Yuan and Jinhua Du and Jinxin Liu and Junkai Zhi and Junwen Duan and Kaiyue Zhou and Kangjian Wei and Ke Wang and Keyun Luo and Laiqiang Zhang and Leigang Sha and Liang Xu and Lindong Wu and Lintao Ding and Lu Chen and Minghao Li and Nianyi Lin and Pan Ta and Qiang Zou and Rongjun Song and Ruiqi Yang and Shangqing Tu and Shangtong Yang and Shaoxiang Wu and Shengyan Zhang and Shijie Li and Shuang Li and Shuyi Fan and Wei Qin and Wei Tian and Weining Zhang and Wenbo Yu and Wenjie Liang and Xiang Kuang and Xiangmeng Cheng and Xiangyang Li and Xiaoquan Yan and Xiaowei Hu and Xiaoying Ling and Xing Fan and Xingye Xia and Xinyuan Zhang and Xinze Zhang and Xirui Pan and Xu Zou and Xunkai Zhang and Yadi Liu and Yandong Wu and Yanfu Li and Yidong Wang and Yifan Zhu and Yijun Tan and Yilin Zhou and Yiming Pan and Ying Zhang and Yinpei Su and Yipeng Geng and Yong Yan and Yonglin Tan and Yuean Bi and Yuhan Shen and Yuhao Yang and Yujiang Li and Yunan Liu and Yunqing Wang and Yuntao Li and Yurong Wu and Yutao Zhang and Yuxi Duan and Yuxuan Zhang and Zezhen Liu and Zhengtao Jiang and Zhenhe Yan and Zheyu Zhang and Zhixiang Wei and Zhuo Chen and Zhuoer Feng and Zijun Yao and Ziwei Chai and Ziyuan Wang and Zuzhou Zhang and Bin Xu and Minlie Huang and Hongning Wang and Juanzi Li and Yuxiao Dong and Jie Tang},
      year={2026},
      eprint={2602.15763},
      archivePrefix={arXiv},
      primaryClass={cs.LG},
      url={https://arxiv.org/abs/2602.15763}, 
}

@misc{yong2026independentsafetyevaluationkimi,
      title={An Independent Safety Evaluation of Kimi K2.5}, 
      author={Zheng-Xin Yong and Parv Mahajan and Andy Wang and Ida Caspary and Yernat Yestekov and Zora Che and Mosh Levy and Elle Najt and Dennis Murphy and Prashant Kulkarni and Lev McKinney and Kei Nishimura-Gasparian and Ram Potham and Aengus Lynch and Michael L. Chen},
      year={2026},
      eprint={2604.03121},
      archivePrefix={arXiv},
      primaryClass={cs.CR},
      url={https://arxiv.org/abs/2604.03121}, 
}

@misc{google_gemini,
  author = {{Google AI}},
  title  = {Gemini: A family of multimodal AI models},
  year   = {2024},
  howpublished = {\url{https://ai.google.dev/gemini}},
  note   = {Accessed: YYYY-MM-DD (e.g., 2025-05-06). Specific version evaluated: Gemini-3.1-Pro-Preview}
}

@inproceedings{DBLP:conf/nips/GorenGE24,
  author       = {Shani Goren and
                  Ido Galil and
                  Ran El{-}Yaniv},
  editor       = {Amir Globersons and
                  Lester Mackey and
                  Danielle Belgrave and
                  Angela Fan and
                  Ulrich Paquet and
                  Jakub M. Tomczak and
                  Cheng Zhang},
  title        = {Hierarchical Selective Classification},
  booktitle    = {Advances in Neural Information Processing Systems 38: Annual Conference
                  on Neural Information Processing Systems 2024, NeurIPS 2024, Vancouver,
                  BC, Canada, December 10 - 15, 2024},
  year         = {2024},
  url          = {http://papers.nips.cc/paper\_files/paper/2024/hash/c8b100b376a7b338c84801b699935098-Abstract-Conference.html},
  bibsource    = {dblp computer science bibliography, https://dblp.org}
}

@inproceedings{DBLP:conf/nips/RenS023,
  author       = {Zhiyuan Ren and
                  Yiyang Su and
                  Xiaoming Liu},
  editor       = {Alice Oh and
                  Tristan Naumann and
                  Amir Globerson and
                  Kate Saenko and
                  Moritz Hardt and
                  Sergey Levine},
  title        = {ChatGPT-Powered Hierarchical Comparisons for Image Classification},
  booktitle    = {Advances in Neural Information Processing Systems 36: Annual Conference
                  on Neural Information Processing Systems 2023, NeurIPS 2023, New Orleans,
                  LA, USA, December 10 - 16, 2023},
  year         = {2023},
  url          = {http://papers.nips.cc/paper\_files/paper/2023/hash/dc81297c791bb989deade65c6bd8c1d8-Abstract-Conference.html},
  bibsource    = {dblp computer science bibliography, https://dblp.org}
}

@inproceedings{DBLP:conf/acl/BaiLMLLLXZWHWJL23,
  author       = {Haoli Bai and
                  Zhiguang Liu and
                  Xiaojun Meng and
                  Wentao Li and
                  Shuang Liu and
                  Yifeng Luo and
                  Nian Xie and
                  Rongfu Zheng and
                  Liangwei Wang and
                  Lu Hou and
                  Jiansheng Wei and
                  Xin Jiang and
                  Qun Liu},
  editor       = {Anna Rogers and
                  Jordan L. Boyd{-}Graber and
                  Naoaki Okazaki},
  title        = {Wukong-Reader: Multi-modal Pre-training for Fine-grained Visual Document
                  Understanding},
  booktitle    = {Proceedings of the 61st Annual Meeting of the Association for Computational
                  Linguistics (Volume 1: Long Papers), {ACL} 2023, Toronto, Canada,
                  July 9-14, 2023},
  pages        = {13386--13401},
  publisher    = {Association for Computational Linguistics},
  year         = {2023},
  url          = {https://doi.org/10.18653/v1/2023.acl-long.748},
  doi          = {10.18653/V1/2023.ACL-LONG.748},
  bibsource    = {dblp computer science bibliography, https://dblp.org}
}

@inproceedings{DBLP:conf/acl/TuGCT23,
  author       = {Yi Tu and
                  Ya Guo and
                  Huan Chen and
                  Jinyang Tang},
  editor       = {Anna Rogers and
                  Jordan L. Boyd{-}Graber and
                  Naoaki Okazaki},
  title        = {LayoutMask: Enhance Text-Layout Interaction in Multi-modal Pre-training
                  for Document Understanding},
  booktitle    = {Proceedings of the 61st Annual Meeting of the Association for Computational
                  Linguistics (Volume 1: Long Papers), {ACL} 2023, Toronto, Canada,
                  July 9-14, 2023},
  pages        = {15200--15212},
  publisher    = {Association for Computational Linguistics},
  year         = {2023},
  url          = {https://doi.org/10.18653/v1/2023.acl-long.847},
  doi          = {10.18653/V1/2023.ACL-LONG.847},
  bibsource    = {dblp computer science bibliography, https://dblp.org}
}

@misc{minimax2026m27,
  title        = {{MiniMax M2.7: Self-Evolving Models Driving Productivity Innovation through Technical Breakthroughs}},
  author       = {{MiniMax}},
  year         = {2026},
  howpublished = {\url{https://www.minimaxi.com/models/text/m27}},
  note         = {Accessed on 2026-05-01},
  key          = {MiniMax M2.7}
}

@article{waples2026opus,
  title     = {Claude Opus 4.7: Anthropic’s New Best (Available) Model},
  author    = {Josef Waples},
  journal   = {DataCamp Blog},
  year      = {2026},
  month     = {apr},
  day       = {16},
  note      = {Updated Apr 17, 2026},
  url       = {https://www.datacamp.com/blog/opus-4-7},
  language  = {english}
}

@INPROCEEDINGS{8260658,
  author={Kowsari, Kamran and Brown, Donald E. and Heidarysafa, Mojtaba and Jafari Meimandi, Kiana and Gerber, Matthew S. and Barnes, Laura E.},
  booktitle={2017 16th IEEE International Conference on Machine Learning and Applications (ICMLA)}, 
  title={HDLTex: Hierarchical Deep Learning for Text Classification}, 
  year={2017},
  volume={},
  number={},
  pages={364-371},
  doi={10.1109/ICMLA.2017.0-134}
  }

@article{SOKOLOVA2009427,
title = {A systematic analysis of performance measures for classification tasks},
journal = {Information Processing \& Management},
volume = {45},
number = {4},
pages = {427-437},
year = {2009},
issn = {0306-4573},
doi = {https://doi.org/10.1016/j.ipm.2009.03.002},
url = {https://www.sciencedirect.com/science/article/pii/S0306457309000259},
author = {Marina Sokolova and Guy Lapalme}
}

@inproceedings{DBLP:conf/cvpr/DengDSLL009,
  author       = {Jia Deng and
                  Wei Dong and
                  Richard Socher and
                  Li{-}Jia Li and
                  Kai Li and
                  Li Fei{-}Fei},
  title        = {ImageNet: {A} large-scale hierarchical image database},
  booktitle    = {2009 {IEEE} Computer Society Conference on Computer Vision and Pattern
                  Recognition {(CVPR} 2009), 20-25 June 2009, Miami, Florida, {USA}},
  pages        = {248--255},
  publisher    = {{IEEE} Computer Society},
  year         = {2009},
  url          = {https://doi.org/10.1109/CVPR.2009.5206848},
  doi          = {10.1109/CVPR.2009.5206848},
  bibsource    = {dblp computer science bibliography, https://dblp.org}
}

@article {PMID:28904516,
	Title = {The Differences and Similarities Between Two-Sample \textit{T}-Test and Paired \textit{T}-Test},
	Author = {Xu, Manfei and Fralick, Drew and Zheng, Julia Z and Wang, Bokai and Tu, Xin M and Feng, Changyong},
	DOI = {10.11919/j.issn.1002-0829.217070},
	Number = {3},
	Volume = {29},
	Month = {June},
	Year = {2017},
	Journal = {Shanghai archives of psychiatry},
	ISSN = {1002-0829},
	Pages = {184--188},
	URL = {https://europepmc.org/articles/PMC5579465},
}

\newpage
\appendix
\section{Appendices}
\subsection{LLM usage}
\label{LLM usage}
\textit{Data Construction Pipeline.}
To facilitate both training and evaluation, we implement a systematic data construction pipeline that converts raw document-annotations pairs into structured instruction samples. The pipeline is implemented in Python using \textit{pandas} for data manipulation and supports multiple input formats (CSV, Excel, JSON, Parquet, Pickle). Random seeds are fixed for reproducibility. The pipeline is supported by a direct prediction (DP) strategy:  models are presented with the complete label taxonomy and select the most relevant category based on document content. This setting evaluates model's intrinsic understanding of both document semantics and  hierarchical label space.

\textit{Prompt Template.}
The complete prompt template used in our experiments is shown in Figure~\ref{fig:prompt_template}. The prompt consists of three sections: (1) a role specification and task description, (2) a flattened hierarchical list of candidate categories with indentation indicating parent-child relationships, and (3) the document input containing OCR text, image page placeholders, and optional filename.

\begin{figure}[htbp]
\centering
\begin{minipage}{0.9\columnwidth}
\begin{verbatim}
# Role & Task
You are a document classification expert.
Analyze the document content and select the most appropriate document type 
Code from the candidate categories.
Output only the Code, and refrain from producing any other characters.

# Candidate Categories
- INVOICE (Invoice)
-   SALES_INVOICE (Sales Invoice)
-   PURCHASE_INVOICE (Purchase Invoice)
- CONTRACT (Contract)
-   SERVICE_CONTRACT (Service Contract)
-   EMPLOYMENT_CONTRACT (Employment Contract)

# Input
Document OCR Text: {ocr_text}
Document Page Images:
<image_1>
<image_2>
File Name: {file_name}
\end{verbatim}
\end{minipage}
\caption{The prompt template used for direct prediction (DP) strategy. The candidate categories are displayed with indentation to reflect hierarchical relationships. Each \texttt{<image\_i>} placeholder is replaced by a special image token (e.g., \texttt{<image>}) depending on the underlying multi-modal model.}
\label{fig:prompt_template}
\end{figure}

\textit{Training and Evaluation Protocol.} The constructed dataset can be used in two primary scenarios: Supervised Fine-Tuning (SFT) and Zero-shot / Few-shot Evaluation. In SFT, Each sample's instruction serves as the user prompt, and the output serves as the expected assistant response. Models are optimized using standard cross-entropy loss on the output tokens. In the  scenario of Zero-shot / Few-shot Evaluation,  the instruction is fed to models without gradient updates, and the generated output is compared against the ground-truth code. Exact-match accuracy serves as the primary metric, supplemented by hierarchical metrics that consider partial credit for ancestor or descendant predictions. For multi-modal implementations, the image field can be resolved to actual image tensors by replacing placeholders with visual tokens before being processed by a vision-language model.

\subsection{Hardware Resource Configuration}
\label{Hardware Resource Configuration}
All inference experiments on open-weight models are conducted on servers equipped with NVIDIA A800 GPUs (80 GB memory each). We use VLLM as the inference framework with BF16 precision. For the Qwen3.5-35B-A3B model, we adopt multi-GPU tensor parallelism on a single node. For all other models (e.g., Qwen2.5-VL-7B, Qwen3-VL-8B, Qwen3.5-9B, Qwen3.5-27B), we deploy multiple single-GPU instances on a single node for parallel inference.


\end{document}